\documentclass[conference]{IEEEtran}
\IEEEoverridecommandlockouts
\usepackage{cite}
\usepackage{amsmath,amssymb,amsfonts}
\usepackage{algorithmic}
\usepackage{graphicx}
\usepackage{textcomp}
\usepackage{xcolor}
\begin{document}

\title{Temporal and Spatial Reservoir Ensembling Techniques for Liquid State Machines\\
}

%\author{\IEEEauthorblockN{Anonymous Authors}}

\author{\IEEEauthorblockN{Anmol Biswas}
\IEEEauthorblockA{\textit{EE Dept. } \textit{IIT Bombay}\\
Mumbai, India \\
194076019@iitb.ac.in}
\and
\IEEEauthorblockN{Sharvari Ashok Medhe}
\IEEEauthorblockA{\textit{EE Dept.} \textit{IIT Bombay}\\
Mumbai, India \\
20d070073@iitb.ac.in}
\and
\IEEEauthorblockN{Raghav Singhal}
\IEEEauthorblockA{\textit{EE Dept.} \textit{IIT Bombay}\\
Mumbai, India \\
19d070049@iitb.ac.in}
\and
\IEEEauthorblockN{Udayan Ganguly}
\IEEEauthorblockA{\textit{EE Dept.} \textit{IIT Bombay}\\
Mumbai, India \\
udayan@ee.iitb.ac.in}
}

\maketitle

\footnote{This work was presented at the International Conference on Neuromorphic Systems (ICONS) 2024, and has been accepted for inclusion in the conference proceedings}

\begin{abstract}
Reservoir computing (RC), which is an umbrella term for a class of computational methods such as Echo State Networks (ESN) and Liquid State Machines (LSM) can be thought of as describing a generic method to perform pattern recognition and temporal analysis with any system having non-linear dynamics. This is enabled by the fact that Reservoir Computing is a shallow network model with only Input, Reservoir, and Readout layers and the input and reservoir weights need not be learned (only the readout layer is trained). LSM is a special case of Reservoir computing inspired by the organization of neurons in the brain and generally refers to spike-based Reservoir computing approaches. LSMs have been successfully used to showcase decent performance on some neuromorphic vision and speech datasets but a common problem associated with LSMs is that since the model is more-or-less fixed, the main way to improve the performance is by scaling up the Reservoir size, but that only gives diminishing rewards despite a tremendous increase in model size and computation. In this paper, we propose two approaches for effectively ensembling LSM models -  Multi-Length Scale Reservoir Ensemble (MuLRE) and Temporal Excitation Partitioned Reservoir Ensemble (TEPRE) and benchmark them on Neuromorphic-MNIST (N-MNIST), Spiking Heidelberg Digits (SHD), and DVSGesture datasets, which are standard neuromorphic benchmarks. We achieve 98.1\% test accuracy on N-MNIST with a 3600-neuron LSM model which is higher than any prior LSM-based approach and 77.8\% test accuracy on the SHD dataset which is on par with a standard Recurrent Spiking Neural Network trained by Backprop Through Time (BPTT). We also propose receptive field-based input weights to the Reservoir to work alongside the Multi-Length Scale Reservoir ensemble model for vision tasks. Thus, we introduce effective means of scaling up the performance of LSM models and evaluate them against relevant neuromorphic benchmarks
\end{abstract}


\section{Introduction}
Reservoir Computing \cite{reservoirComp} is a general class of shallow network models that consist of Input, Reservoir, and output/classifier. In these models, only the output classifier weights are learned, the Input-Reservoir connections and the recurrent Reservoir connections are not learned, and the computational/representational power is derived from the complex dynamics of the Reservoir itself. It is this simplicity of the Reservoir Computing model that allows it to convert potentially any system with complex dynamic behavior into a powerful computing unit. Liquid State Machines (LSM) \cite{maassLSM} is a special case of Reservoir computing, where the Reservoir is composed of a population of spiking neurons with random and recurrent connections. Thus it is also a Spiking Neural Network (SNN) \cite{maassSNN} and can be evaluated on neuromorphic benchmarks.
In the absence of training (by design) in the Input and Reservoir layers, there is no universally accepted method to optimize the LSM setup. For example, NALSM \cite{nalsm} and P-Critical \cite{pcritical} propose modulated Spike time-Dependent Plasticity (STDP) to optimize the Input and Reservoir connections, \cite{statespaceLSM} proposes a state-space model for fast evaluation of any given Reservoir connections matrix, MAdapter\cite{madapter} proposes a kind of transfer learning for Input layer hyperparameters that have been optimized on any given dataset and SpiLinC \cite{spilinc} proposes an ensemble LSM model. MAdapter\cite{madapter} also shows significant performance improvement by a process called time-partitioning but does not comment on the physical realizability of time-partitioning. In this paper, we adapt time-partitioning as a form of temporal ensemble model, called Temporal Excitation Partitioned Reservoir Ensemble (TEPRE) where each Reservoir in the ensemble only "sees" the input for a fixed fraction of the total presentation time of the sample. We also propose a Multi-Length Scale Reservoir Ensemble (MuLRE) model where each Reservoir in the ensemble has different underlying probability distributions for connecting neurons in the Reservoir, thus creating greater diversity in the representations generated by the ensemble model. Finally, we also propose a receptive-field-based input weight matrix for the LSM to preserve the spatial orderedness of vision input in the Reservoir as well. We evaluate all of our methods on standard neuromorphic benchmark datasets, i.e. NMNIST \cite{nmnist}, SHD \cite{shd}, and DVSGesture \cite{dvsgesture}. Code available at \textit{https://github.com/SNNalgo/snntorch-LSM}

\begin{figure*}[htp]
\centerline{\includegraphics[width=0.6\linewidth]{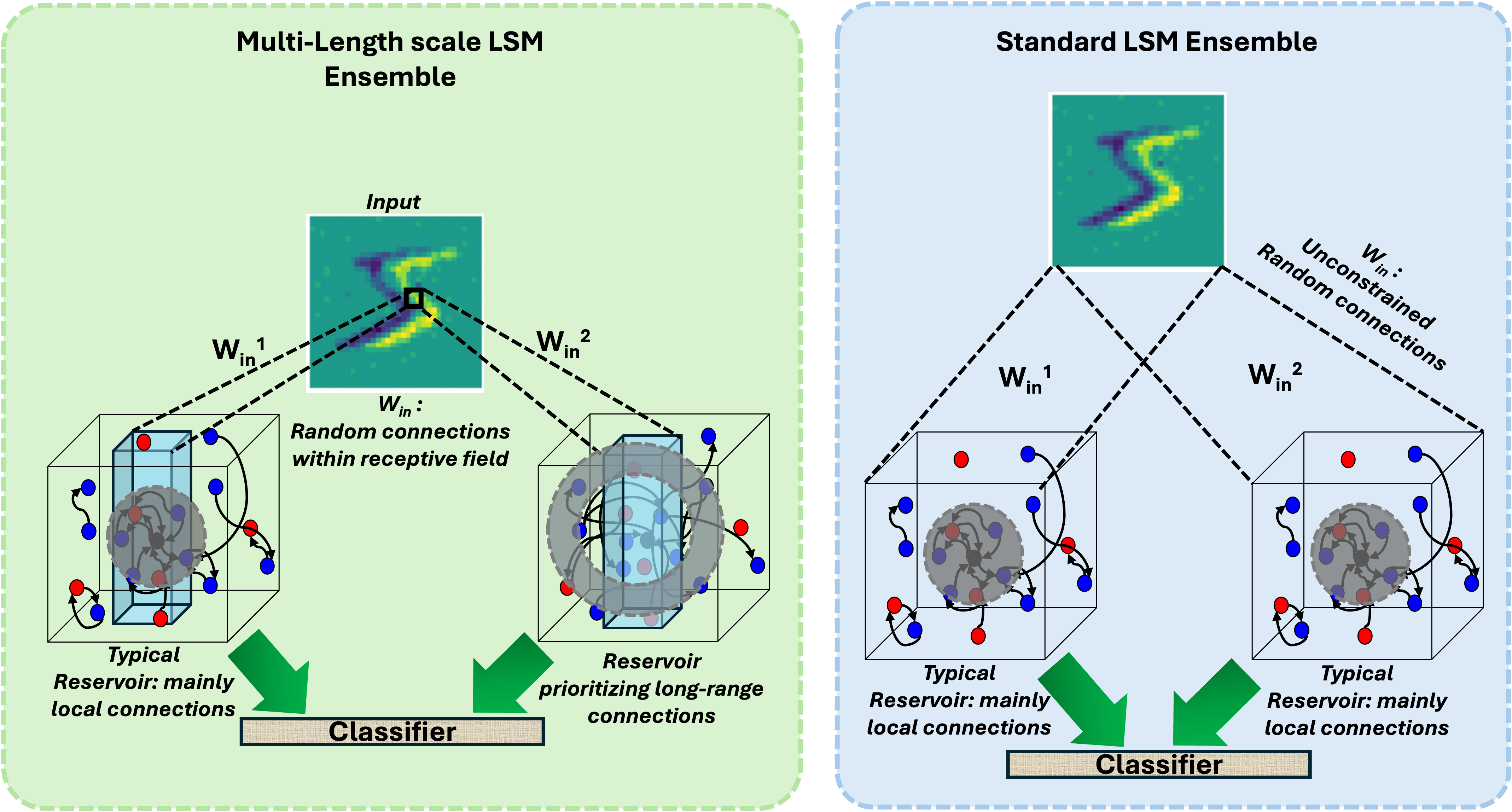}}
\caption{Multi-Length Scale Reservoir Ensemble (MuLRE) compared against a standard LSM ensemble setup. MuLRE must use Receptive Field-based Input connections}
\label{distance_ens}
\end{figure*}

\section{Proposed Method}

\subsection{Reservoir Neuron Model}
The Reservoir neurons are Leaky-Integrate and Fire neurons described as follows \cite{loihi}:
\begin{equation}
    dv_i/dt = -v_i(t)/\tau_v + u_i(t) - \theta\sigma_i(t)
    \label{eq1}
\end{equation}

\begin{equation}
    u_i(t) = \sum_{j\neq i}w_{j,i}(\sigma_j*\alpha_u)(t)
    \label{eq2}
\end{equation}

\begin{equation}
    \sigma_i(t) = \sum_t\delta(t-t^k_i)
    \label{eq3}
\end{equation}

Where $v_i(t)$ is the membrane potential of the $i^{th}$ Reservoir neuron, $u_i(t)$ is the corresponding synaptic current, $\tau_v$ is the decay time constant of the membrane potential, $\sigma_i(t)$ is the spike-train of the $i^{th}$ Reservoir neuron and $\theta$ is the spiking threshold

In eq. \ref{eq2}, $w_{j,i}$ is the weight connecting neuron $j$ to neuron $i$ where neuron $i$ is always a Reservoir neuron, but neuron $j$ can be either a different Reservoir neuron or an input neuron, and $\alpha_u(t)=(1/\tau_u)(e^{-t/\tau_u})H(t)$ is the synapse current kernel, where $H(t)$ is the \textit{Heaviside} function and $\tau_u$ is the time constant of the synaptic current

For all of our simulations, we have set $\theta=20$ and used $\Delta t=1$ as the time-step

\subsection{Input-Reservoir}
\subsubsection{Standard Input}
For \textbf{Standard Input}, the Input is simply converted to a flat vector, and then an equal number of positive and negative connections are made from each input neuron to the Reservoir. The number of connections to be made is decided by the hyperparameter: \textbf{input connection density}, as described in \cite{madapter} and \cite{nalsm}. In this paper, we use the MAdapter\cite{madapter} method to set the Input layer hyperparameters \textbf{input weight} and \textbf{input connection density}

\subsubsection{Receptive Field-based Input}
For \textbf{Receptive Field-based Input}, the process of making connections from any Input neuron to the Reservoir does not change compared to \textbf{Standard Input}, but the pool of Reservoir neurons that an Input neuron can connect to changes. Instead of potentially being able to connect to any neuron in the Reservoir, now the input neuron can connect only to neurons within a narrow square window around the $(x,y)$ co-ordinates in the Reservoir (the Reservoir is to be considered as a 3-D grid of neurons) that correspond to the $(x, y)$ co-ordinates of the input neuron in question. The window size is a hyperparameter and is set to $5$ or $6$ for our experiments. This restriction allows the spatial order of the visual input to be preserved even inside the Reservoir. This method of connecting Input to the Reservoir is visualized in Fig. \ref{distance_ens}
%\begin{figure}[htbp]
%\centerline{\includegraphics[width=\columnwidth]{receptive_field_input.png}}
%\caption{Receptive Field-based Input}
%\label{receptive_in}
%\end{figure}

\begin{figure}[htbp]
\centerline{\includegraphics[width=0.8\linewidth]{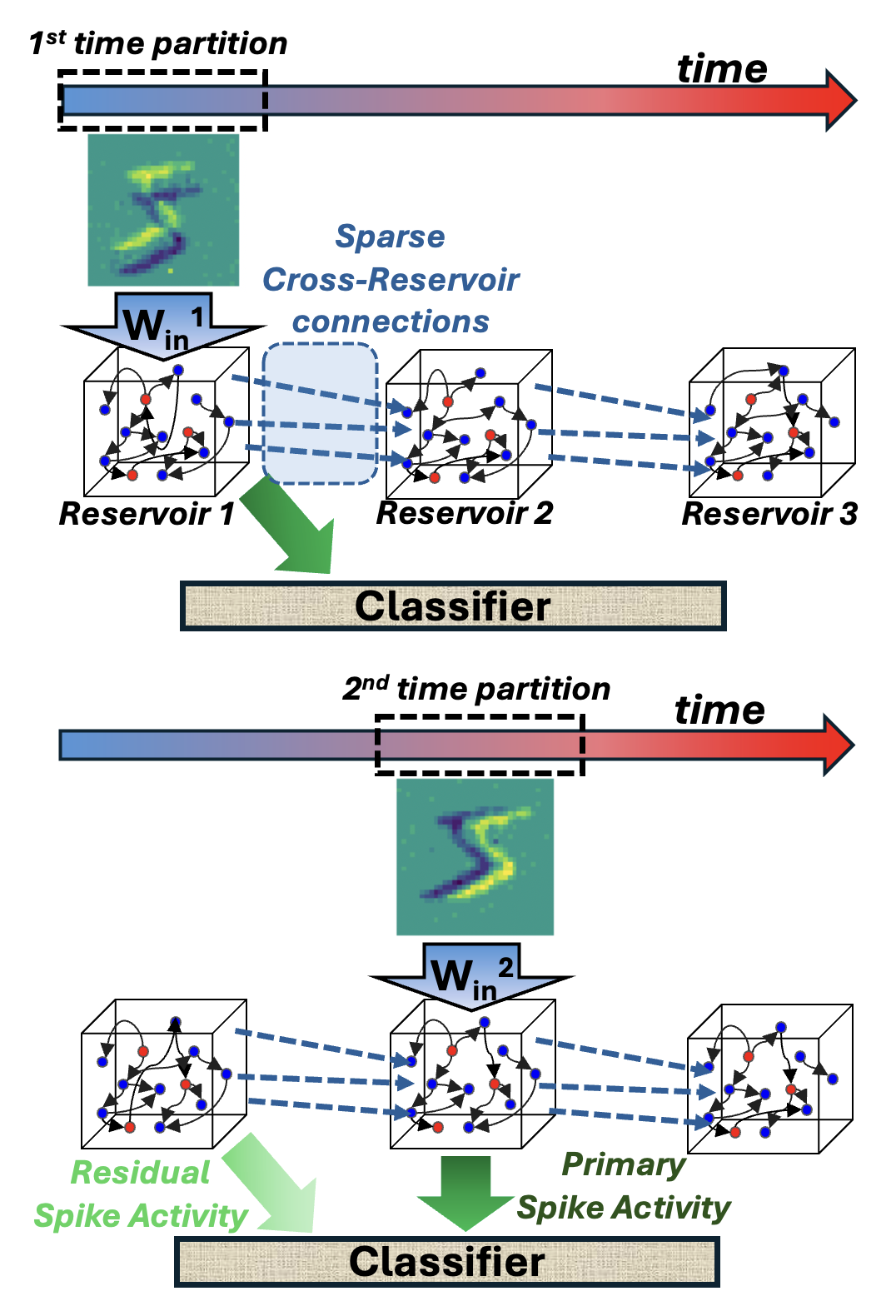}}
\caption{Temporal Excitation Partitioned Reservoir Ensemble (TEPRE)}
\label{temporal_ens}
\end{figure}

\subsection{Reservoir Connections}
The Reservoir is conceptualized as a 3-dimensional grid of spiking neurons, with dimensions $N_x$, $N_y$ and $N_z$. Half of the Reservoir neurons are excitatory and the other half are inhibitory. Generally, for a local probabilistic model of connectivity \cite{localprob}, connections between neuron $i$ and neuron $j$ within the Reservoir as made with the probability distribution:
\begin{equation}
    Prob(i,j) = C exp(-(D(i,j)/\lambda)^2)
    \label{eq4}
\end{equation}
Where $D(i,j)$ is the Euclidean distance between the two neurons. By default, this biases the Reservoir to prefer making short-distance connections between neurons. For the Multi-Length Scale Reservoir Ensemble model, we use a slight variation of this distribution function by introducing a distance term $d$ as follows:
\begin{equation}
    Prob(i,j) = C exp(-((D(i,j)-d)/\lambda)^2)
    \label{eq5}
\end{equation}
This distribution biases connections for which the neurons are approximately $d$ distance apart within the Reservoir
The weight value in the Reservoir, i.e. $w_{lsm}$ from excitatory neurons and $-w_{lsm}$ from inhibitory neurons is set to $1$ and the values of $C$ to be used depend on the types of neurons $i$ and $j$ as described in \cite{nalsm}. For \textit{excitatory-excitatory} connections, $C=0.2$, for \textit{excitatory-inhibitory} connections, $C=0.1$, for \textit{inhibitory-excitatory} connections, $C=0.05$, and for \textit{inhibitory-inhibitory} connections, $C=0.3$

\subsection{Ensemble Models}

\subsubsection{Multi-Length Scale Reservoir Ensemble}
For the \textbf{Multi-Length Scale Reservoir Ensemble model} (MuLRE), we simply use different values of $d$ depending on the number of reservoirs to be used for the ensemble. For 2-Reservoir ensemble, we use $d=\{0, 5\}$ and for 3-Reservoir ensemble, we use $d=\{0, 4, 6\}$. The MuLRE is always used in conjunction with Receptive Field-based Input. On the N-MNIST dataset, we observed best performance after preprocessing the incoming frames by a bank of 18 Gabor filters\cite{gabor}. The ensemble setup is visualized in Fig. \ref{distance_ens}

\subsubsection{Temporal Excitation Partitioned Reservoir Ensemble}
For the \textbf{Temporal Excitation Partitioned Reservoir Ensemble} (TEPRE) we consider a number of smaller partition Reservoirs equal to the number of time partitions and schedule the input in such a way that it is received by one partition Reservoir within each time partition. The individual partition Reservoirs are interconnected with very sparse inhibitory connections that try to prevent successive partition Reservoirs from generating the same or highly correlated output. The operation of the TEPRE method is demonstrated in Fig. \ref{temporal_ens}. For TEPRE experiments we use Standard Input only.

\subsection{Classifier}
In all of our experiments, we perform the final classification using a simple linear classifier ($linear\_model$ from $sklearn$ library)\cite{sklearn}

\section{Experiments and Results}
\subsection{N-MNIST}
\begin{figure}[htbp]
\centerline{\includegraphics[width=\columnwidth]{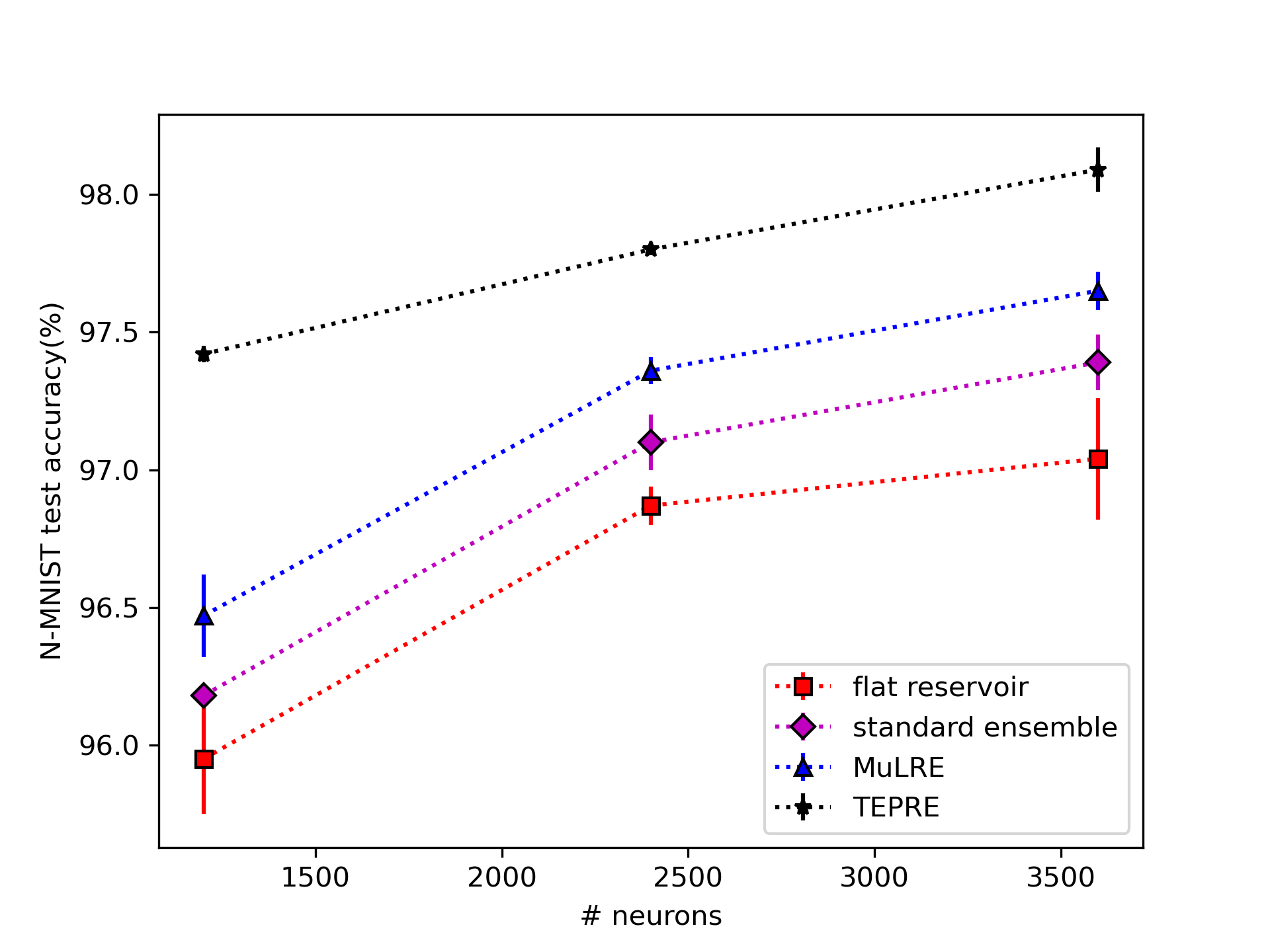}}
\caption{Comparison of the different ensemble methods}
\label{ensemble_comp}
\end{figure}

We primarily test all of our proposed models on the N-MNIST dataset because it suits all the models being developed. We use $\tau_u = \tau_v = 16$ for this dataset and test with different settings of number of partitions and ensemble configurations. Fig. \ref{ensemble_comp} compares the two proposed methods against a vanilla ensemble LSM model and simple LSM (not ensemble) with an equal number of neurons. We find that the MuLRE significantly improves upon the vanilla ensemble LSM model, but is outperformed by the TEPRE model which achieves a test accuracy of 98.1\% with a 3600-neuron reservoir and 3 partitions

\begin{figure}[htbp]
\centerline{\includegraphics[width=\columnwidth]{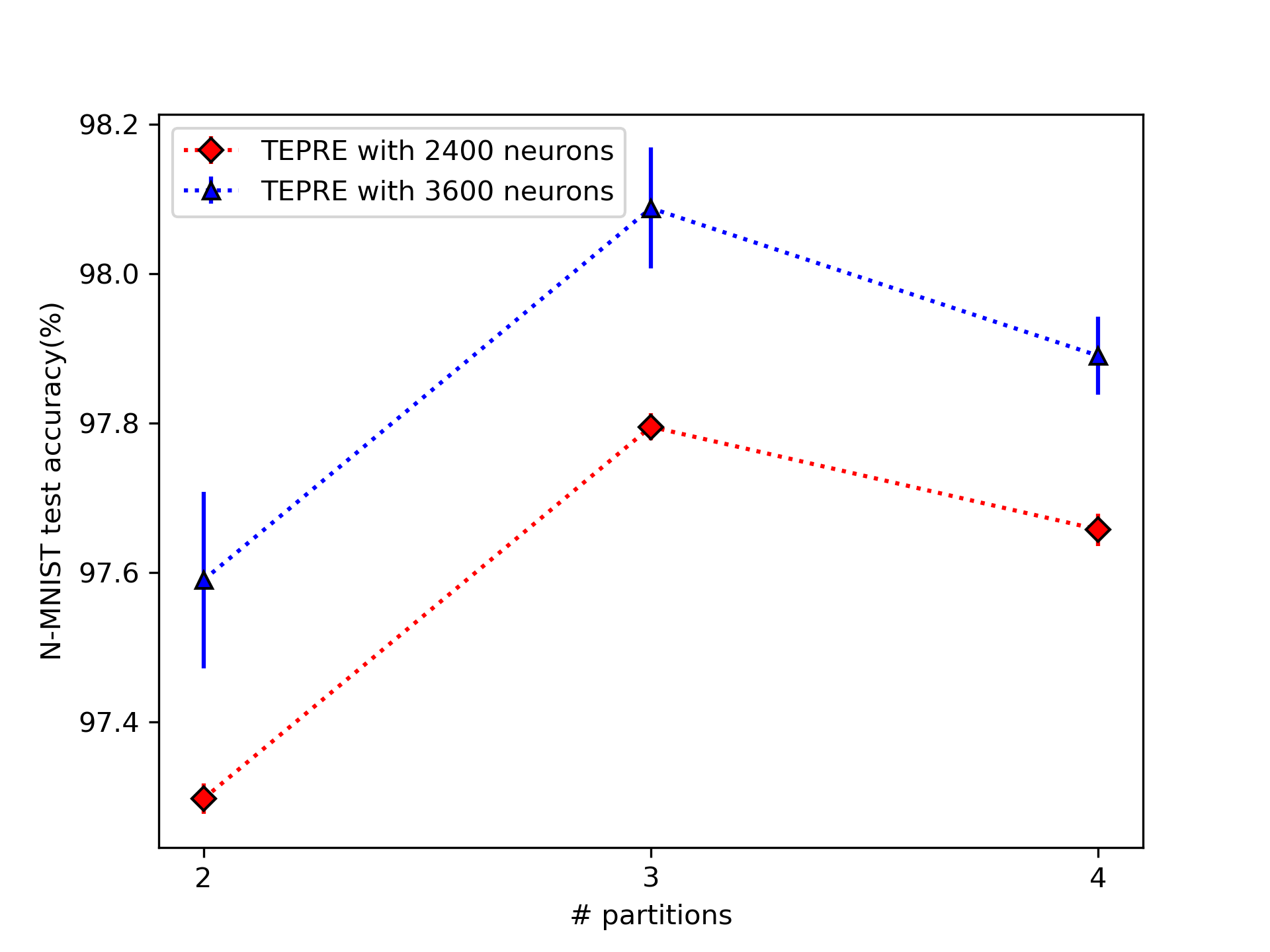}}
\caption{Performance vs number of partitions}
\label{partitions_comp}
\end{figure}

Fig. \ref{partitions_comp} compares the effect of increasing the number of partitions on the test accuracy and we find that 3 partitions is the optimal setting. This is expected considering that the N-MNIST dataset programs 3 temporally separate motions on the digit images. This indicates in the direction that the optimal number of partitions is likely to be decided by the underlying features of the data

\subsection{SHD}
The Spiking Heidelberg Digits\cite{shd} is a speech benchmark for Spiking Neural Networks. As the data consists of 1-D vector inputs at every time step, it is unsuitable for the MuLRE model. The dataset is read with time\_window=1000 in Tonic\cite{tonic} and we tested it with the TEPRE model and achieved a maximum test accuracy of $77.8\%$ using $(\tau_v, \tau_u) = (40, 20)$, $N_x=N_y=10, N_z=30$ and 6 partitions. This is comparable to the performance of a Recurrent Spiking Neural Network (RSNN) trained by BPTT\cite{shd}.

\subsection{DVSGesture}
DVSGesture\cite{dvsgesture} dataset is a vision dataset where the task is to identify 11 gestures from their DVS camera recordings. Each input frame read with time\_window=20000 in Tonic\cite{tonic} is 128x128x2 pixels, which we scale down to 64x64x2 before sending it to a 20x20x10 ($N_x$x$N_y$x$N_z$) Reservoir using Receptive field-based input connections. Unfortunately, the DVSGesture dataset only contains ~1000 training examples and it suffers from overfitting if any of the approaches discussed in this paper are applied. Under these restrictions, the maximum test accuracy achieved is $85.2\%$ using $(\tau_v, \tau_u) = (5, 10)$, $N_x=N_y=20, N_z=10$ and 1 partition.

\subsection{Comparison}
\begin{table}[htbp]
\caption{}
\begin{tabular}{|p{0.3\linewidth}|p{0.25\linewidth}|p{0.15\linewidth}|p{0.1\linewidth}|}
\hline
\textbf{Model} & \textbf{Learning} & \textbf{Accuracy} & \textbf{training params}\\
\hline
\multicolumn{4}{|c|}{\textbf{Dataset: N-MNIST}}\\
\hline
\textbf{NALSM8000}\cite{nalsm} & \textbf{astro-STDP} + last layer\newline Gradient descent & 97.51\% & 80K\\
\hline
\textbf{SLAYER}\cite{slayer} & \textbf{Backpropagation} & \textbf{98.89}\% & 1.41M\\
\hline
\textbf{This work}, \newline 3600-neuron TEPRE,\newline partition=3 & last layer\newline Gradient descent & \textbf{98.1}\% & 36K\\
\hline
\textbf{This work}, \newline 3600-neuron Multi-Length Scale Reservoir Ensemble & last layer\newline Gradient descent & 97.65\% & 36k\\
\hline
\multicolumn{4}{|c|}{\textbf{Dataset: SHD}}\\
\hline
\textbf{This work}, \newline 3000-neuron TEPRE,\newline partition=6 & last layer\newline Gradient descent & \textbf{77.8}\% & 30K\\
\hline
\textbf{RSNN}\cite{shd} & \textbf{BPTT} & \textbf{79.9}\% & 1.78M\\
\hline
\multicolumn{4}{|c|}{\textbf{Dataset: DVSGesture}}\\
\hline
\textbf{This work}, \newline 4000-neuron lsm with\newline Standard input & last layer\newline Gradient descent & 82.8\% & 40K\\
\hline
\textbf{This work}, \newline 4000-neuron lsm with\newline Receptive Field-based input & last layer\newline Gradient descent & 85.2\% & 40K\\
\hline
\textbf{Spiking CNN with 5 Conv layers and 2 Fully connected\newline layers} & \textbf{Surrogate Gradient}\cite{snntorch} & \textbf{95.34}\% & 5.38M\\
\hline
\end{tabular}
\label{table1}
\end{table}

\section{Conclusion}
In this paper, we propose two LSM ensemble models and one new LSM input approach that greatly improve the performance and scaling of the LSM approach to temporal classification, both generally, and specifically for vision inputs. With these LSM ensemble models, LSM-based methods are approaching the standard learning-based baselines for N-MNIST and SHD benchmarks. This furthers the generalization of LSM and RC-based approaches. Future works could be finding suitable data augmentation techniques to improve the performance in the DVSGesture dataset for which the ensemble models overfit or implementing more complex neuron and synapse models, (including STDP) in the Reservoir to enhance its information representation properties

\bibliographystyle{ieeetr}
\bibliography{references}

\end{document}